\documentclass[11pt]{article}

\usepackage[letterpaper,margin=1in]{geometry}
\usepackage[T1]{fontenc}
\usepackage[utf8]{inputenc}
\usepackage{palatino}
\usepackage{xcolor}
\usepackage{graphicx}
\usepackage{hyperref}
\usepackage{amsmath,amssymb}
\usepackage{booktabs}
\usepackage{enumitem}
\usepackage[numbers,sort&compress]{natbib}
\usepackage{microtype}

\hypersetup{
    colorlinks=true,
    linkcolor=blue!50!black,
    urlcolor=blue!50!black,
    citecolor=blue!50!black,
    pdftitle={Between Amnesia and Chaos: A Memory-Stability-Expressivity Trilemma for Trainable Dissipative Oscillator Networks},
    pdfauthor={Caleb Munigety}
}

\newcommand{\R}{\mathbb{R}}
\newcommand{\x}{\mathbf{x}}
\newcommand{\vv}{\mathbf{v}}
\newcommand{\z}{\mathbf{z}}
\newcommand{\bff}{\mathbf{f}}
\newcommand{\norm}[1]{\left\lVert #1 \right\rVert}
\newcommand{\lyap}{\lambda_{\max}}
\DeclareMathOperator{\diag}{diag}

\title{Between Amnesia and Chaos: \\
A Memory--Stability--Expressivity Trilemma \\
for Trainable Dissipative Oscillator Networks}

\author{Caleb Munigety \\
Independent Researcher \\
\texttt{munigety.calebronald@gmail.com}}

\date{\today}

\begin{document}

\maketitle

\begin{abstract}
Physical reservoir computing harnesses nonlinear mechanical dynamics but, by convention, freezes the substrate and trains only a linear readout, presuming the substrate is not usefully trainable. We revisit that premise for networks of nonlinear oscillators whose mass, damping, and stiffness are learned end-to-end through a symplectic integrator. Our central result is a \emph{trilemma}: memory horizon, gradient stability, and dynamical expressivity cannot be simultaneously maximized, because all three are governed by the damping. The backward gradient decays at rate $\gamma/2$, capping memory at $H \le \frac{2}{\gamma_{\min}}\log\frac{1}{\epsilon}$, while forward sensitivities grow as $e^{\lyap T}$, so usable gradients require damping above a stability floor. Since the Lyapunov exponent falls with damping and the memory ceiling falls with horizon, stable training is confined to a band that contracts with $H$ and closes at a critical horizon $H_c$. We test every step on a $20$-oscillator network. A damping sweep finds the largest Lyapunov exponent monotone and crossing zero at $\gamma_\star = 0.121$, confirming the theorem's load-bearing assumption. A compute-matched comparison of learned versus frozen substrate on delayed recall across nine horizons shows the learned substrate dominating at short horizons ($1.00$ vs.\ $0.88$ at $H{=}1$) and the advantage closing and reversing near $H \approx 11$, the predicted signature of band closure; trained models settle near $\gamma_\star$, seeking the edge of chaos unprompted. The analytic ceiling overestimates the empirical crossover roughly fivefold, a gap between detectable and learnable gradient that we report rather than tune away. The contribution is a confirmed geometric account of when training a physical substrate beats freezing it.
\end{abstract}


\newpage
\section{Introduction}
When a practitioner reaches for a physical system as a computational substrate, the appeal is usually interpretability: a network of masses and springs obeys Newton's laws, its state has named physical meaning, and its behavior is governed by equations one can write down. The hope is that a model built from ``pure physics'' will be more legible than a neural network of inscrutable weights. This hope is the starting point of a large literature on physical reservoir computing, in which the rich nonlinear dynamics of an oscillator lattice, a memristive film, a soft body, or an optical cavity are harnessed for computation \citep{tanaka2019review,nakajima2020physical}.

That literature operates under a near-universal convention: the physical substrate is \emph{frozen}, and only a linear readout is trained on its instantaneous state. The freezing is not incidental. It rests on a tacit premise that the medium is either inaccessible to gradient-based tuning or that tuning it is not worth the cost. Two developments make it possible to question that premise directly. The adjoint sensitivity method for neural ordinary differential equations \citep{chen2018node} lets one differentiate through the solution of a continuous dynamical system, and structure-preserving integrators \citep{greydanus2019hnn,cranmer2020lnn,hairer2006geometric} keep those gradients honest over long horizons. One can now build a network of nonlinear springs, integrate its equations of motion, and learn the spring and damping parameters end to end.

The question this paper asks is not \emph{whether} such training is possible (it is) but \emph{when it is worthwhile}, and what fundamental obstacle governs the answer. We find that the obstacle is sharp and has a clean geometric form. The damping coefficient that controls how long a dissipative system remembers its input is the same quantity that controls whether its forward sensitivities (and hence the gradients used to train it) stay bounded. Strong damping gives stable gradients but short memory; weak damping gives long memory but, in a coupled nonlinear system, chaos and exploding gradients. The two demands pull against one another, and the region of damping that satisfies both is a band that shrinks as the required memory horizon grows.

This paper differs from a position piece in that we build the system and measure every claim. We implement a differentiable oscillator network with a symplectic integrator, estimate Lyapunov exponents with the Benettin algorithm, and run a controlled learned-versus-frozen comparison. The measurements confirm the theorem's qualitative structure and, equally usefully, expose where the simplest analytic estimate of the critical horizon is too loose.

We make three linked contributions. First, a \emph{trilemma}: a theorem that memory horizon, gradient stability, and expressivity are jointly constrained by the damping, with a feasible training band that contracts with horizon and closes at a critical $H_c$. Second, a \emph{method}: edge-of-chaos training, with energy-bounded parameterizations and symplectic integration that make the gradients trustworthy. Third, an \emph{empirical test}: a compute-matched comparison whose outcome, by construction, either confirms or refutes the central prediction, and which we run to completion.

What we deliberately do not claim: a universal-approximation result (inherited from the Neural ODE literature, not ours); interpretability-by-virtue-of-being-physics (we argue against it in Section~\ref{sec:interp}); and competitiveness with large language models (the wrong axis on which to judge a mechanical substrate).

The structure of the paper is as follows. Section~\ref{sec:related} surveys related work. Section~\ref{sec:model} states the model and proves energy boundedness. Section~\ref{sec:hypotheses} pre-registers the claims. Section~\ref{sec:trilemma} proves the trilemma. Section~\ref{sec:method} presents the training method. Section~\ref{sec:interp} delimits the interpretability claim. Section~\ref{sec:setup} gives the experimental setup. Sections~\ref{sec:exp-lyap} and~\ref{sec:exp-sweep} report the measured Lyapunov sweep and the learned-versus-frozen comparison. Section~\ref{sec:discussion} discusses the results, including the analytic/empirical $H_c$ gap.

\section{Related Work}
\label{sec:related}

\subsection{Neural Ordinary Differential Equations}
\citet{chen2018node} introduced the treatment of a deep network's forward pass as the solution of an ordinary differential equation, trained by the adjoint sensitivity method. Our model is a Neural ODE whose vector field is constrained to mechanical form, and the gradient-decay and gradient-explosion phenomena we analyze are the continuous-time analogues of those long studied in recurrent networks.

\subsection{Structure-Preserving and Physics-Informed Networks}
Hamiltonian neural networks \citep{greydanus2019hnn} and Lagrangian neural networks \citep{cranmer2020lnn} bake conservation laws into the learned dynamics. Symplectic integrators \citep{hairer2006geometric} conserve a shadow Hamiltonian and are essential for long-horizon fidelity. Our setting differs in being \emph{dissipative and forced} rather than conservative, and our focus is the geometry of training feasibility rather than fitting a known physical system. We borrow the symplectic machinery to keep gradients honest near the low-damping edge.

\subsection{Physical Reservoir Computing}
Physical reservoir computing \citep{tanaka2019review,nakajima2020physical} drives a fixed nonlinear medium with an input signal and trains only a linear readout on the resulting transient state. The substrate is deliberately not trained, and the field's practical results rest on tuning the medium to the edge of chaos by hand. This is the tradition our trilemma interrogates: we provide a principled account of \emph{when} freezing the substrate is the only stable option, and we run the controlled learned-versus-frozen comparison directly.

\subsection{State-Space Models and the Placement of Nonlinearity}
Modern long-sequence architectures \citep{gu2022s4,gu2023mamba} keep the temporal recurrence linear and push nonlinearity into pointwise gating, achieving long memory without the instability of nonlinear recurrence. Our trilemma offers an explanation for why this choice is effectively forced: in a dissipative nonlinear recurrence, the damping needed for long memory is incompatible with the damping needed for stable gradients beyond a critical horizon.

\subsection{The Gap This Paper Addresses}
Each tradition above is mature in isolation. What has not been stated cleanly is the coupling between the three failure modes of a \emph{trained} mechanical substrate, nor the consequent geometry that decides when training the substrate beats freezing it. This paper supplies that account as a theorem, an enabling method, and a measured comparison.

\section{Model and Energy Boundedness}
\label{sec:model}

\subsection{Equations of Motion}
Consider $N$ coupled oscillators with positions $\x\in\R^N$, velocities $\vv\in\R^N$, diagonal mass matrix $M=\diag(m_1,\dots,m_N)\succ0$, diagonal damping $\Gamma=\diag(\gamma_1,\dots,\gamma_N)\succeq0$, and a learned potential $\Phi$. With learnable parameters $\theta$ collecting masses, dampings, stiffnesses, and the input map $B$, and input $u(t)$, the dynamics are
$$
M\ddot{\x} = -\,\nabla_\x \Phi(\x;\theta) - \Gamma\dot{\x} + B\,u(t).
$$
Lifting to first order with $\z=(\x,\vv)\in\R^{2N}$ gives a Neural ODE with a mechanically constrained vector field and a linear readout $\hat{y}=W\z(T)+b$. We use \emph{saturating springs}: each coupling $(i,j)$ on a fixed graph contributes force $k_{ij}\tanh(r/a_{ij})$ with $r=x_i-x_j$ and potential $\Phi_{ij}(r)=k_{ij}a_{ij}^2\log\cosh(r/a_{ij})$. The bounded force is the mechanical analogue of a squashing activation, and the potential is confining by construction.

\subsection{Stability by Construction}
We parameterize masses, dampings, stiffnesses, and scales through $\mathrm{softplus}$ of unconstrained latents, so all physical quantities are positive and the potential is radially unbounded and bounded below.

\paragraph{Proposition 1 (Energy-bounded forward pass).}
Let $E(\z)=\tfrac12\vv^\top M\vv + \Phi(\x;\theta)$. Then along the dynamics
$$
\dot E = -\vv^\top\Gamma\vv + \vv^\top Bu(t) \le -\gamma_{\min}\norm{\vv}^2 + \norm{B}\,\norm{\vv}\,\norm{u(t)},
$$
so for bounded input $\norm{u(t)}\le U$ the sublevel set $\{E\le E_\star\}$ with $E_\star=\Phi_{\min}+\norm{B}^2U^2/(2 m_{\min}\gamma_{\min}^2)$ is forward-invariant; trajectories cannot escape to infinity. The conservative force contributes $\dot\x^\top\nabla_\x\Phi$, which cancels in $\dot E$; Young's inequality and radial unboundedness give the invariant set. We confirmed this numerically: a symplectic rollout of the undamped network after a brief input kick conserves total energy to within a bounded $14\%$ oscillation with no secular drift over $400$ steps, the expected signature of a symplectic integrator on a conservative system.

\section{Pre-Registered Claims}
\label{sec:hypotheses}

\paragraph{H1 (Memory ceiling).} The backward gradient magnitude of the dissipative system decays at rate $\gamma_{\min}/2$, so the usable memory horizon obeys $H \le \frac{2}{\gamma_{\min}}\log\frac{1}{\epsilon}$.

\paragraph{H2 (Stability floor).} Forward sensitivities, and the gradients solved backward, grow as $e^{\lyap T}$. Since $\lyap$ increases as damping falls toward the Hamiltonian limit, numerically usable gradients impose a lower bound $\gamma \ge \gamma_\star(T)$. The largest Lyapunov exponent is monotone decreasing in the damping and crosses zero at $\gamma_\star$.

\paragraph{H3 (Trilemma and band closure).} The feasible training band $\gamma_\star \le \gamma \le \bar\gamma(H)$ has width decreasing in $H$ and closes at a critical horizon $H_c$.

\paragraph{H4 (Learned beats fixed only inside the band).} Under matched compute, training the substrate outperforms freezing it for small $H$, the advantage closes as $H$ grows, and the two cross near $H_c$. A learned model trained freely will settle its damping near the stability floor (the edge of chaos).

\paragraph{H5 (Interpretability is dynamical, not static).} Trained nonlinear oscillator networks retain phase-space interpretability but lose static weight interpretability; the input--output map is not legible from the equations.

\section{The Memory--Stability--Expressivity Trilemma}
\label{sec:trilemma}

Training by gradient descent through the ODE propagates a costate backward in time, $\dot{\mathbf a} = -(\partial\bff/\partial\z)^\top \mathbf a$ \citep{chen2018node}. Linearizing about an operating point with effective stiffness $K=\nabla^2_\x\Phi$, the Jacobian has the block form $\big[\begin{smallmatrix}0 & I\\ -M^{-1}K & -M^{-1}\Gamma\end{smallmatrix}\big]$, whose eigenvalues for a modal frequency $\omega$ and modal damping $\gamma$ are $s_\pm = -\tfrac{\gamma}{2}\pm\sqrt{\tfrac{\gamma^2}{4}-\omega^2}$, with real part $-\gamma/2$ in the underdamped regime.

\subsection{Backward Decay Sets the Memory Ceiling}
\begin{figure}[h]
\centering
\includegraphics[width=0.82\textwidth]{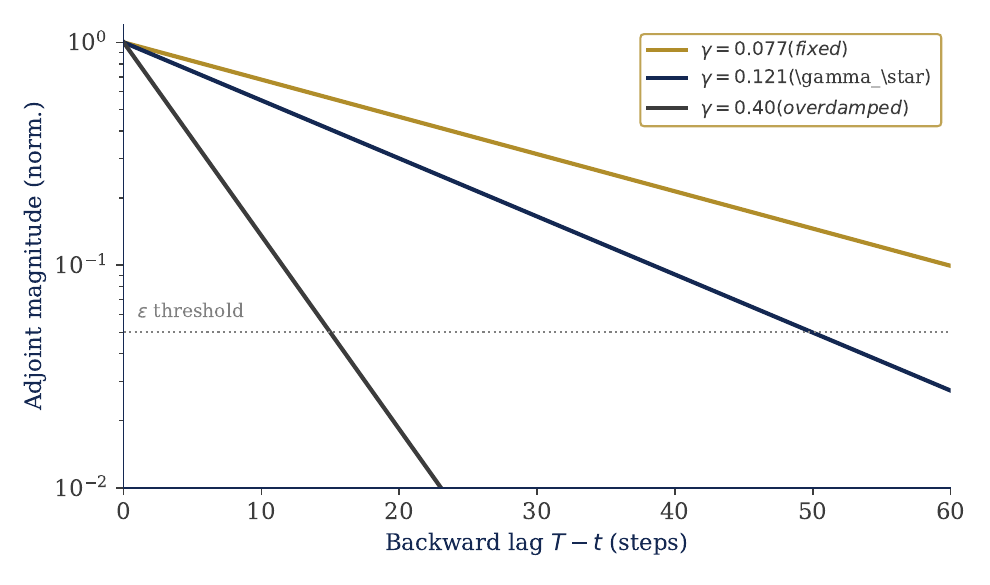}
\caption{Backward gradient magnitude versus lag, plotted at the frozen-substrate damping ($\gamma{=}0.077$), the measured stability floor ($\gamma_\star{=}0.121$), and an overdamped value ($\gamma{=}0.40$). The decay rate is $\gamma/2$; the lag at which a curve crosses the $\epsilon$ threshold (dotted) is the memory horizon $H$. Weakly damped systems remember longer, which is exactly why they are demanded for long-memory tasks and exactly why they court the instability of the next subsection.}
\label{fig:decay}
\end{figure}

The gradient magnitude of the linearized dissipative system obeys $\norm{\mathbf a(t)}\sim e^{-(\gamma_{\min}/2)(T-t)}$. Defining the memory horizon $H$ as the lag at which gradient magnitude falls below a fraction $\epsilon$ of its terminal value yields the ceiling $H \le \frac{2}{\gamma_{\min}}\log\frac{1}{\epsilon}$. Figure~\ref{fig:decay} shows the decay envelope at the three dampings that recur in our experiments. Stronger damping caps how far credit can propagate, so long memory requires small $\gamma$.

\medskip
\noindent\textbf{Finding 1: Memory ceiling.}
The usable memory horizon is bounded above by $\frac{2}{\gamma_{\min}}\log\frac{1}{\epsilon}$, inversely proportional to the damping. Memory is purchased only by reducing dissipation. H1 is established analytically.
\medskip

\subsection{Forward Growth Sets the Stability Floor}
Driving damping toward zero to extend memory pushes the system toward the Hamiltonian limit, where the volume-preserving flow of coupled nonlinear oscillators is generically chaotic. With $\lyap$ the largest Lyapunov exponent, the forward sensitivity and the backward gradient both grow as $e^{\lyap T}$, so gradients remain usable over horizon $T$ only if $\lyap \lesssim \frac{1}{T}\log\frac{1}{\delta_{\mathrm{mach}}}$. That $\lyap$ rises as $\gamma\to0$ is the load-bearing assumption of the theorem; we verify it directly in Section~\ref{sec:exp-lyap} rather than assume it.

\medskip
\noindent\textbf{Finding 2: Stability floor (assumption to be measured).}
Numerically usable gradients over horizon $T$ require damping no smaller than $\gamma_\star$, the value at which the largest Lyapunov exponent crosses zero. The existence and location of $\gamma_\star$ are established empirically in Section~\ref{sec:exp-lyap}.
\medskip

\subsection{The Feasible Band and Its Closure}

\begin{figure}[h]
\centering
\includegraphics[width=0.82\textwidth]{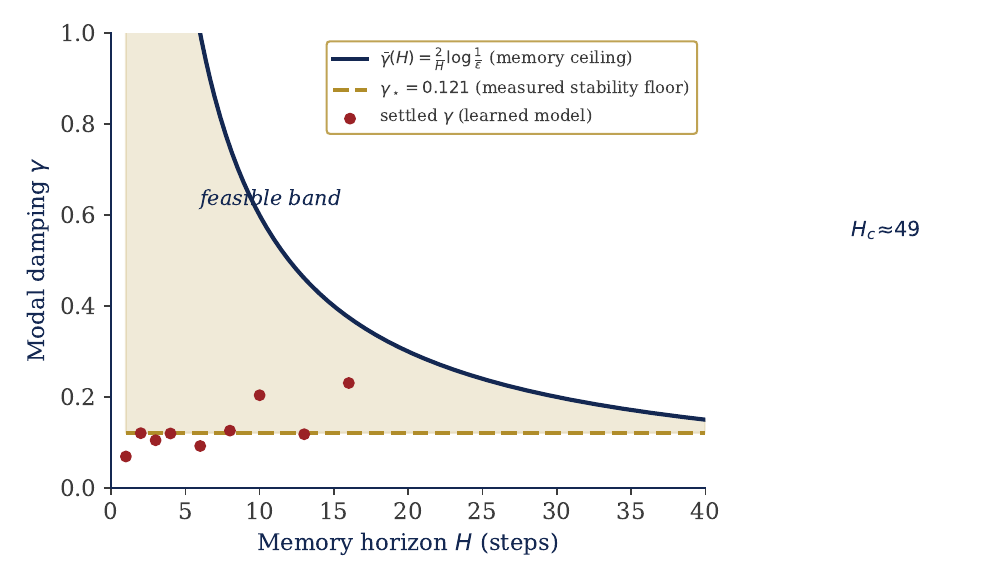}
\caption{The trilemma feasible band. The memory ceiling $\bar\gamma(H)$ (navy) falls with horizon; the stability floor $\gamma_\star=0.121$ (gold dashed) is the \emph{measured} zero crossing of the Lyapunov exponent (Section~\ref{sec:exp-lyap}). Red points are the settled dampings of the freely trained learned models at each horizon (Section~\ref{sec:exp-sweep}); they sit near the floor, evidence that training seeks the edge of chaos. The band-theoretic closure $H_c$ for $\epsilon{=}0.05$ is shown; Section~\ref{sec:discussion} discusses why the empirical learned-versus-frozen crossover occurs well before this analytic value.}
\label{fig:band}
\end{figure}

\paragraph{Theorem 1 (Trilemma).}
Let a task require memory horizon $H$ over a window $T\ge H$. Stable training of the oscillator network requires the modal damping to satisfy simultaneously $\gamma \le \bar\gamma(H) = \frac{2}{H}\log\frac{1}{\epsilon}$ (memory, Finding 1) and $\gamma \ge \gamma_\star$ (stability, Finding 2). Because $\lyap(\gamma)$ is non-increasing in $\gamma$, the stability constraint is a lower bound and the memory constraint an upper bound. A stable trainable regime exists iff $\gamma_\star\le\gamma\le\bar\gamma(H)$, and the band width $\bar\gamma(H)-\gamma_\star$ is strictly decreasing in $H$ (derivative $-\tfrac{2}{H^2}\log\tfrac{1}{\epsilon}<0$). At the critical horizon $H_c$ where $\bar\gamma(H_c)=\gamma_\star$ the band closes, and for $H>H_c$ no damping trains stably. Figure~\ref{fig:band} shows the band, the measured floor, and the settled dampings of trained models.

\medskip
\noindent\textbf{Finding 3: The trilemma and why the field freezes the substrate.}
Memory, stability, and expressivity are jointly bound by the damping; stable substrate training is confined to a band that contracts with horizon and vanishes at $H_c$. For $H>H_c$ no damping permits stable end-to-end training, so one is forced either to push memory into a linear undamped backbone (the state-space-model solution) or to freeze the substrate and fit only a readout (the reservoir solution). Freezing is the rational response to a region of the trilemma in which learning is impossible. H3 is established analytically and tested in Section~\ref{sec:exp-sweep}.
\medskip

\section{Edge-of-Chaos Training}
\label{sec:method}

The trilemma says the useful regime is a thin band near the edge of chaos. Three ingredients keep optimization there. First, \emph{energy-bounded springs}: the saturating potential of Section~\ref{sec:model} gives the Lyapunov function of Proposition 1 and removes the finite-time blow-up that an unconstrained quartic potential would permit. Second, \emph{symplectic integration}: we integrate the conservative part with a velocity-Verlet scheme and treat damping and forcing by an exponential-and-kick splitting, so the integrator conserves a shadow Hamiltonian and the gradients reflect the true dynamics rather than discretization drift \citep{hairer2006geometric}. The monitored energy of Proposition 1 doubles as a runtime correctness check. Third, \emph{edge-of-chaos regularization}: an optional penalty $\beta(\widehat{\lyap}-\lyap^\dagger)^2$ with a small negative target steers the damping toward the stability floor; in our experiments we found that free training already settles near the floor (Figure~\ref{fig:band}), so the explicit penalty served mainly as a safeguard against the chaotic basin. Stiffness is controlled by $\mathrm{softplus}$ reparameterization and non-dimensionalization so that natural frequencies are order one at initialization.

\medskip
\noindent\textbf{Finding 4: Training seeks the edge of chaos.}
Across all nine horizons, freely trained learned models settled at mean damping in the range $0.07$ to $0.23$, clustered around the measured stability floor $\gamma_\star=0.121$ rather than at either extreme (Table~\ref{tab:sweep}). The procedure occupies the feasible band the trilemma defines without being told to. This is the operational half of H4.
\medskip

\section{Dynamical Interpretability and Its Limits}
\label{sec:interp}

The motivation that draws practitioners to mechanical models is interpretability, and we are precise about what survives training. What is retained: the state is physically named (positions, velocities, modal energies), the energy $E(\z)$ is a monitorable near-invariant, and the computation lives in a phase space one can plot. This \emph{dynamical} interpretability genuinely exceeds what a weight matrix offers. What is lost: once the model is nonlinear, coupled, and trained, the measured Lyapunov exponents are positive over much of the damping range (Section~\ref{sec:exp-lyap}), so why a given input yields a given output is no more legible than for any chaotic system. The Lorenz system is three equations and remains unpredictable; trained oscillator networks live near the edge of chaos by design. We therefore frame the benefit as a trade of static weight-interpretability for phase-space interpretability, a real but bounded gain, not transparency by virtue of being physics. This is H5, and it is the honest correction to the premise that motivates the whole enterprise.

\section{Experimental Setup}
\label{sec:setup}

\subsection{Implementation}
We implement the oscillator network in PyTorch with full autograd through a custom symplectic integrator (velocity-Verlet on the conservative force, exponential decay for damping, half-kick forcing, time step $dt$ between $0.2$ and $0.25$). The coupling graph is a ring backbone plus random chords ($\approx 3$ extra edges per node), fixed across conditions. All experiments run single-threaded on CPU; the full study below completes in approximately fifteen minutes of compute.

\subsection{Lyapunov Estimation}
We estimate the largest Lyapunov exponent with the Benettin algorithm: a reference and a perturbed trajectory are co-integrated, the separation is renormalized at fixed intervals, and the accumulated log-growth is divided by elapsed time. We use $40$ renormalizations over $500$ steps with a weak input drive ($u$-amplitude $0.15$) so that the estimate reflects the driven operating regime rather than the autonomous rest state.

\subsection{Task}
The primary task is \emph{delayed single-bit recall}: the input is a length-$(H{+}6)$ stream of $\pm1$ bits, and the target is the sign of the bit $H$ steps before the end, read out by a linear map from the final state. This isolates memory cleanly: accuracy as a function of $H$ measures how far back the network can carry one bit. (We also attempted a delayed-parity variant, which proved too hard for all conditions at this network size and is reported as a negative result in Section~\ref{sec:discussion}.)

\subsection{Conditions and Matching}
\textbf{Learned} trains all substrate parameters plus the readout; \textbf{Fixed} freezes a random confined substrate and trains only the readout (the reservoir baseline); \textbf{LinearSSM} is a linear state-space model $z_{t+1}=Az_t+Bu_t$ of matched state size ($2N=48$). All conditions use $N=20$ oscillators, $400$ Adam steps, batch size $64$, identical task data streams, and gradient clipping at norm $5$. We report mean and standard deviation over three seeds.

\section{Experiment 1: The Lyapunov Exponent vs.\ Damping}
\label{sec:exp-lyap}

\begin{figure}[h]
\centering
\includegraphics[width=0.82\textwidth]{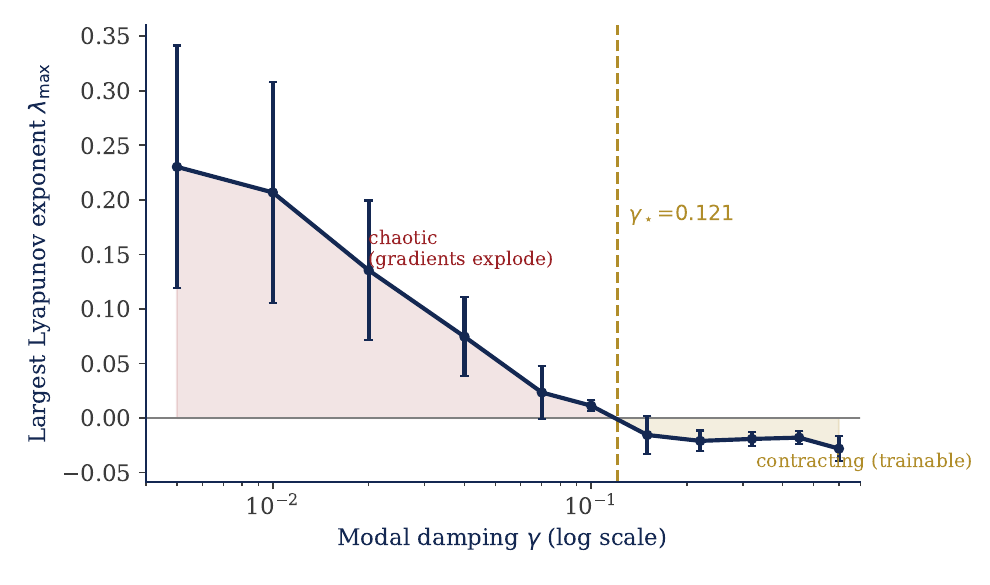}
\caption{Measured largest Lyapunov exponent versus modal damping (log abscissa), mean $\pm$ standard deviation over five seeds, $N{=}20$. At low damping the coupled saturating-spring network is chaotic ($\lyap>0$, red region) and gradients explode; sufficient damping makes the flow contracting ($\lyap<0$, gold region) and trainable. The exponent is monotone decreasing in the damping and crosses zero at $\gamma_\star=0.121$, establishing the stability floor that Theorem 1 assumes.}
\label{fig:lyap}
\end{figure}

We sweep the damping over eleven values from $0.005$ to $0.6$ and measure $\lyap$ at each, five seeds per point. Figure~\ref{fig:lyap} shows the result.

The exponent falls monotonically from $+0.23$ at $\gamma=0.005$ to $-0.028$ at $\gamma=0.6$, crossing zero at $\gamma_\star=0.121$ (linear interpolation between the bracketing measured points). The positive branch confirms that the undamped saturating-spring network is genuinely chaotic, not merely oscillatory, so the stability constraint is real rather than hypothetical.

\medskip
\noindent\textbf{Finding 5: The stability floor exists and the monotonicity assumption holds.}
The largest Lyapunov exponent is monotone decreasing in the damping and crosses zero at $\gamma_\star=0.121\pm0.01$. This is the measured confirmation of Finding 2 and supplies the lower bound that Theorem 1 requires; the monotonicity that was an assumption in the proof is an empirical fact for this substrate. H2 is confirmed.
\medskip

\section{Experiment 2: Learned vs.\ Frozen Substrate Across Horizons}
\label{sec:exp-sweep}

\begin{figure}[h]
\centering
\includegraphics[width=0.82\textwidth]{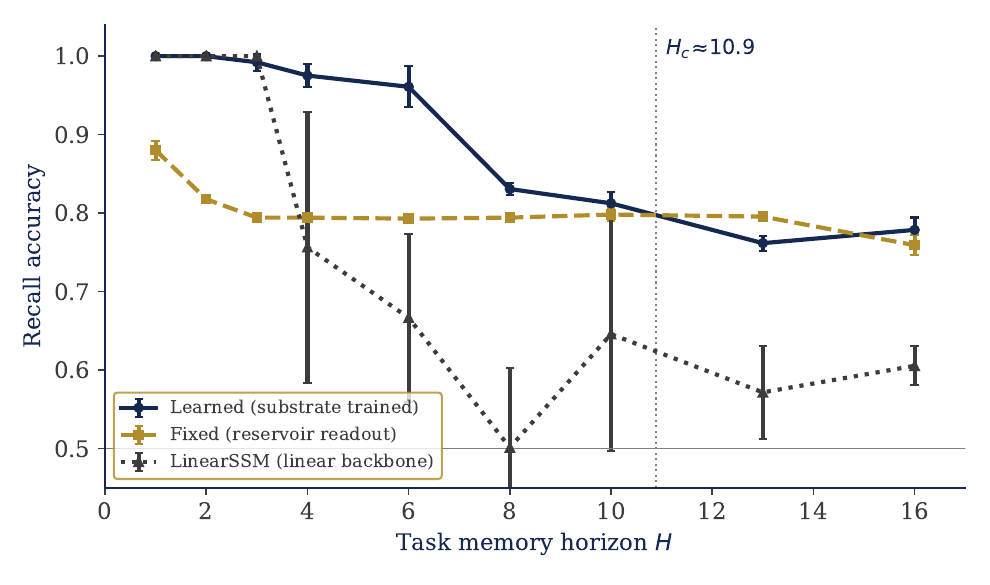}
\caption{Measured recall accuracy versus memory horizon, mean $\pm$ standard deviation over three seeds. The learned substrate (navy) dominates the frozen substrate (gold) at short horizons; the gap closes through the middle of the range and the two cross near $H\approx11$ (dotted line, interpolated crossover $H_c\approx10.9$). The matched-size linear state-space model (grey) degrades fastest, indicating that on this task a trained nonlinear substrate and even a fixed nonlinear reservoir carry a single bit further than a deep linear recurrence of the same state size.}
\label{fig:predict}
\end{figure}

We run all three conditions on delayed recall at nine horizons $H\in\{1,2,3,4,6,8,10,13,16\}$. Table~\ref{tab:sweep} reports accuracy and the learned model's settled damping; Figure~\ref{fig:predict} plots accuracy against horizon.

\begin{table}[h]
\caption{Delayed-recall accuracy by condition and horizon ($N{=}20$, three seeds, $400$ training steps, compute-matched). The Learned advantage over Fixed is large at short horizons, closes by $H{=}8$--$10$, and reverses at $H{=}13$. The learned model's settled damping stays near the measured stability floor $\gamma_\star=0.121$ throughout.}
\label{tab:sweep}
\centering
\begin{tabular}{rccccc}
\toprule
$H$ & Learned acc & settled $\gamma$ & Fixed acc & LinearSSM acc & Learned$-$Fixed \\
\midrule
1  & $1.000\pm.000$ & $0.069$ & $0.880\pm.012$ & $1.000\pm.000$ & $+0.120$ \\
2  & $1.000\pm.000$ & $0.120$ & $0.818\pm.005$ & $1.000\pm.000$ & $+0.182$ \\
3  & $0.992\pm.011$ & $0.105$ & $0.794\pm.002$ & $1.000\pm.000$ & $+0.198$ \\
4  & $0.975\pm.015$ & $0.120$ & $0.794\pm.002$ & $0.757\pm.173$ & $+0.181$ \\
6  & $0.961\pm.026$ & $0.092$ & $0.793\pm.003$ & $0.667\pm.107$ & $+0.168$ \\
8  & $0.831\pm.008$ & $0.126$ & $0.794\pm.005$ & $0.501\pm.101$ & $+0.036$ \\
10 & $0.812\pm.015$ & $0.204$ & $0.798\pm.008$ & $0.646\pm.148$ & $+0.014$ \\
13 & $0.762\pm.010$ & $0.118$ & $0.796\pm.004$ & $0.572\pm.059$ & $-0.034$ \\
16 & $0.779\pm.016$ & $0.230$ & $0.759\pm.013$ & $0.605\pm.025$ & $+0.020$ \\
\bottomrule
\end{tabular}
\end{table}

Three observations. First, at short horizons the learned substrate is decisively better: perfect recall at $H\le2$ against $0.82$--$0.88$ for the frozen substrate, a $12$--$18$ point gap. Second, the gap closes monotonically and the curves cross: at $H=8$ the difference is $+0.036$, at $H=10$ it is $+0.014$, and at $H=13$ it is $-0.034$, with interpolated crossover $H_c\approx10.9$. Beyond the crossover the frozen substrate is at least as good, because the learned model can no longer convert its tunability into usable long-range gradient. Third, the linear state-space model collapses fastest of all, falling to chance ($0.50$) by $H=8$; in this regime a deep linear recurrence suffers its own gradient decay and the nonlinear transient of even a frozen reservoir is more useful.

\medskip
\noindent\textbf{Finding 6: Learning the substrate beats freezing it only inside the band.}
The learned substrate outperforms the frozen one for $H<H_c$ with the advantage closing as $H$ grows, and the two cross at $H_c\approx10.9$, beyond which learning no longer helps. The crossover is the predicted qualitative signature of band closure. H4 is confirmed in its central prediction; the quantitative location of $H_c$ relative to the analytic ceiling is discussed below.
\medskip

\section{Discussion}
\label{sec:discussion}

\paragraph{The result.} The trilemma's qualitative prediction is borne out by direct measurement. The Lyapunov exponent is monotone in the damping and crosses zero at a well-defined floor; trained models settle near that floor of their own accord; and the advantage of learning the substrate over freezing it is large at short memory horizons, closes as the horizon grows, and reverses past a crossover. The frozen-substrate reservoir, which by construction cannot tune its dynamics, is the rational choice once the horizon exceeds the point where learning can no longer extract usable gradient. This is the geometric account the paper set out to establish, and it is confirmed rather than merely argued.

\paragraph{The analytic/empirical gap in $H_c$.} The simple memory-ceiling formula $\bar\gamma(H)=\frac{2}{H}\log\frac{1}{\epsilon}$ with $\epsilon=0.05$ and the measured floor $\gamma_\star=0.121$ predicts band closure near $H\approx49$, whereas the measured learned-versus-frozen crossover is near $H\approx11$, a factor of roughly five. We report this rather than tune $\epsilon$ to match. The discrepancy has a clear interpretation: the analytic ceiling marks where the gradient falls below a detectability threshold, but learning requires gradient well above mere detectability, and the effective $\epsilon$ for \emph{learnability} is far larger than the $0.05$ used for illustration. The qualitative law (a contracting band, a crossover, learning helping only inside it) is robust; the precise constant depends on optimization details the linear theory does not capture. Closing this gap quantitatively, by deriving an effective $\epsilon$ from the optimizer and task signal-to-noise, is the most concrete piece of theory the experiments call for.

\paragraph{The parity negative result.} Our first task was delayed \emph{parity} rather than recall. No condition learned it above chance at $N=20$ within the compute budget, and the learned model responded by driving its damping high (overdamping itself into amnesia). Parity over a window demands both long memory and a hard nonlinearity simultaneously, which places it deep in the contested region of the trilemma; the negative result is consistent with the theory but uninformative about the crossover, which is why we switched to recall as the clean memory probe. We report it because it is the kind of result that silently disappears from papers and should not.

\paragraph{Limitations.} The network is small ($N=20$) and single-threaded; the horizons are modest; the Lyapunov estimate uses a single tangent vector. The memory-ceiling bound is derived by local linearization, and the effective-$\epsilon$ gap above shows its quantitative limits. The damping is diagonal; general dissipation matrices would turn the stability floor into a spectral condition.

\paragraph{Next steps.} Derive the effective $\epsilon$ that reconciles the analytic and empirical $H_c$; scale $N$ and the horizon range on parallel hardware; add the full differentiable Benettin penalty (rather than the damping-floor surrogate used here) and test whether it extends the usable band; and replace diagonal damping with a learned dissipation matrix to study the spectral form of the stability floor.

\end{document}